\documentclass[letterpaper, 10 pt, conference]{ieeeconf}  
\IEEEoverridecommandlockouts
\usepackage{array}
\newcolumntype{P}[1]{>{\centering\arraybackslash}p{#1}}
\newcolumntype{M}[1]{>{\centering\arraybackslash}m{#1}}
\usepackage{tikz}
\definecolor{ctop1}{rgb}{0.5,1.0,0.2}
\definecolor{ctop2}{rgb}{0.5,1.0,0.2}
\definecolor{ctop3}{rgb}{0.5,1.0,0.2}
\newcommand{\tikzcircle}[2][red,fill=red]{\tikz[baseline=-0.5ex]\draw[#1,radius=#2] (0,0) circle ;}%
\usepackage{cite}
\usepackage{float}
\usepackage{caption}
\usepackage{graphicx}

\usepackage{hyperref} 

\usepackage{etoolbox}
\makeatletter
\patchcmd{\@makecaption}
  {\scshape}
  {}
  {}
  {}
\patchcmd{\@makecaption}
  {\\}
  {.\ }
  {}
  {}
\patchcmd{\@makecaption}
  {\centering}
  {}
  {}
  {}
\makeatother

\usepackage{environ}         
\usepackage{etoolbox}        
\usepackage{graphicx}        

\newlength{\myl}
\let\origequation=\equation
\let\origendequation=\endequation

\RenewEnviron{equation}{
  \settowidth{\myl}{$\BODY$}                       
  \origequation
  \ifdimcomp{\the\linewidth}{>}{\the\myl}
  {\ensuremath{\BODY}}                             
  {\resizebox{\linewidth}{!}{\ensuremath{\BODY}}}  
  \origendequation
}

\usepackage{array}
\newcolumntype{M}[1]{>{\centering\arraybackslash}m{#1}}
\newcolumntype{N}{@{}m{0pt}@{}}

\usepackage{xcolor}

\title{\LARGE \bf
Commonsense Scene Graph-based Target Localization for Object Search 
}

\author{Wenqi Ge$^{1}$, Chao Tang$^{1}$, Hong Zhang$^{1,*}$ 
\thanks{$^{1}$ Shenzhen Key Laboratory of Robotics and Computer Vision, SUSTech, Shenzhen, China. {\tt \small12232112@mail.sustech.edu.cn}}
\thanks{$^{*}$corresponding author: Hong Zhang {\tt \small hzhang@sustech.edu.cn} }
\thanks{This work was supported by the Shenzhen Key Laboratory of Robotics and Computer Vision (ZDSYS20220330160557001).}
}

\IEEEoverridecommandlockouts
\begin{document}

\maketitle
\thispagestyle{empty}
\pagestyle{empty}

\begin{abstract}

Object search is a fundamental skill for household robots, yet the core problem lies in the robot's ability to locate the target object accurately. The dynamic nature of household environments, characterized by the arbitrary placement of daily objects by users, makes it challenging to perform target localization. To efficiently locate the target object, the robot needs to be equipped with knowledge at both the object and room level. However, existing approaches rely solely on one type of knowledge, leading to unsatisfactory object localization performance and, consequently, inefficient object search processes. To address this problem, we propose a commonsense scene graph-based target localization, CSG-TL, to enhance target object search in the household environment. Given the pre-built map with stationary items, the robot models the room-level spatial knowledge with object-level commonsense knowledge generated by a large language model (LLM) to a commonsense scene graph (CSG), supporting both types of knowledge for CSG-TL. To demonstrate the superiority of CSG-TL on target localization, extensive experiments are performed on the real-world ScanNet dataset and the AI2THOR simulator. Moreover, we have extended CSG-TL to an object search framework, CSG-OS, validated in both simulated and real-world environments. Code and videos are available at \href{https://sites.google.com/view/csg-os}{https://sites.google.com/view/csg-os}.

\end{abstract}

\section{Introduction}

For household robots frequently interacting with human environments, efficiently locating a target object is essential for completing subsequent tasks, such as object searching for downstream manipulation. 
For instance, in the scenario where a user instructs a robot to find a wallet, ideally, the robot should prioritize areas where the wallet is likely to be found, such as on a sofa or a chair, rather than inefficiently searching in unlikely places like a refrigerator.
Considering the intricate interactions between movable targets (objects that users frequently place) and stationary items at the room-level and object-level across diverse indoor settings, it is a challenge to locate the movable targets by their correlation with known stationary items.
To leverage commonsense knowledge of both room-level and object-level for efficient target object localization, similar to human reasoning, we equip the robot with a model that combines the room-level layout and object-level commonsense knowledge. Training this model on real-world scenarios enables the robot to locate targets effectively by utilizing both spatial and contextual insights.

To achieve such a goal, recent efforts focus on leveraging the correlation between target objects and stationary items to enhance target location efficiency, which further benefits object search. Zheng \textit{et al.} \cite{ZhengKicra2022} primarily utilizes object co-occurrence probabilities to frame object search within the context of a partially observable Markov decision process (POMDP). However, this approach depends on the availability of precise object correlations for specific scenes, which is a lack of general object-level knowledge that is often not feasible. Subsequent study \cite{zsos} has explored using graph neural networks (GNNs) to infer these correlations from the Visual Genome \cite{vgdataset} image dataset. Yet, the image dataset can capture only partial views of scenes, which lacks the room-level knowledge that restricts the models' ability to understand complete room-level layouts, which is critical for a robot's comprehensive understanding of its environment.

\begin{figure}[tp]
    \centering
    \includegraphics[width=1\linewidth]{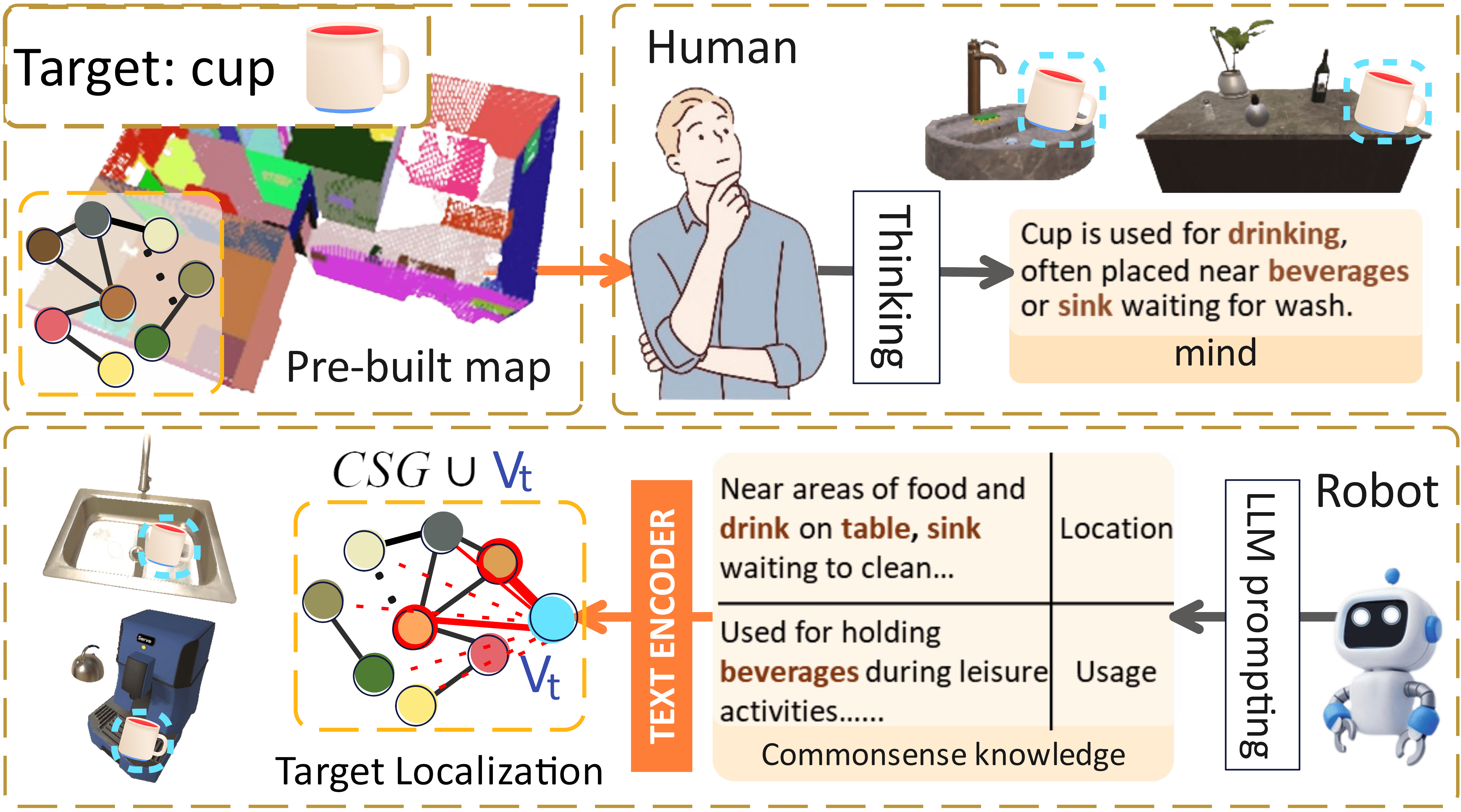}
    \caption{An overview of target object location strategies comparing human room-level and object-level commonsense-inspired reasoning to our CSG-TL method, which integrates commonsense scene graph (CSG) with room-level and object-level knowledge for efficient target localization.}
    \label{fig:sec1-1}
\end{figure}

Concurrently, large language models (LLMs) with vast commonsense knowledge, have shown promise in enhancing robot tasks \cite{dallebot, graspgpt}. Despite their potential, their lack of spatial contextualization hinders directly applying LLMs to target localization in household settings \cite{llmre}, as they struggle to capture the relationships between objects and their surroundings effectively.
To leverage LLMs' strengths and address the limitations of existing works and LLMs, we propose a novel commonsense scene graph-based target localization method, CSG-TL, which integrates object-level commonsense from LLMs with room-level spatial layouts from pre-built maps. 
In contrast to previous works \cite{ZhengKicra2022, zsos} dependent on statistical correlations without object-level commonsense or struggles with capturing room-level object correlations due to limited viewpoints, our model captures both the room-level spatial layouts from pre-built maps and object-level commonsense knowledge obtained from LLMs. This fusion allows our method to utilize the rich interconnections among objects and benefit the human-like reasoning process for target localization.
We integrate CSG-TL into the object search task and contribute to the CSG-OS framework. Specially, in our approach, a robot’s search for a target object in a household environment involves two key phases: target localization (TL) and target object search (OS). During the TL phase, the robot estimates the probable location of the target object by leveraging its correlations at both the room and object levels with stationary items from the pre-built map. Following that, in the OS phase, the robot proceeds to the location predicted by TL to verify the presence of the target object. An illustrative overview of our approach is illustrated in Fig. \ref{fig:sec1-1}.

Our approach, CSG-TL, initiates with creating the commonsense scene graph (CSG), which integrates the room-level spatial relationships of stationary items from pre-built maps with the object-level commonsense knowledge derived from an LLM.
This process allows us to represent a target within the CSG as a node, addressing target localization as a link prediction problem.
Here, the existence of links between the target and other nodes within the CSG infers potential co-occurrences between the target and respective items, thereby implementing the target localization.
CSG-TL is the key component of our proposed object search (CSG-OS) framework, 
after projecting the likelihood of where the target may be located, as determined by CSG-TL, onto the pre-built map. We then cluster these projections to identify candidate navigation points for search. We trained CSG-TL on the ScanNet dataset, which contains over a thousand real-world scenes with labeled object relationships. 
To demonstrate its effectiveness and ability to generalize (zero-shot capability), we tested it on both the ScanNet \cite{scannet} and the AI2THOR environments \cite{ai2thor}, achieving SOTA results. The overall CSG-OS framework, incorporating CSG-TL, also achieves the SOTA performance. Additionally, we tested our CSG-OS framework on the Jackal mobile robot, successfully demonstrating its real-world applicability.

In summary, our key contributions are as follows:
\begin{itemize}
  \item We introduce CSG-TL, an innovative method that merges the knowledge at both room-level from pre-built maps and object-level commonsense obtained from an LLM into a commonsense scene graph (CSG), facilitating superior target localization.
  \item Building on CSG-TL, we propose a commonsense scene graph-based object search (CSG-OS) framework, with the SOTA performance in a simulation study and successful deployment on the real robot.
\end{itemize}

\section{Related Work}
Object search involves a wide range of subproblems and different types of target objects. We consider targets as movable objects (users frequently place) and an environment where the location of stationary items is given as in \cite{ZhengKicra2022,zsos}.

\textbf{Object Search}
Object navigation is traditionally conceptualized as finding a particular category object in an unknown environment \cite{objsearch,searchunknow1}. 
Previous approaches \cite{cows,expobjnav1} focus on building a semantic map or keeping an episodic memory during the search, aiming to enhance efficiency by preventing repeated searches in the same area.
Nonetheless, these approaches take much time to explore, even searching for different targets in the same scene, rendering them inefficient for household robots equipped with a pre-built map. 
To improve search efficiency, maps and stationary items can be integrated. \cite{ZhengKicra2022} employed co-occurrence probabilities between targets and stationary items, framing object search within the context of a partially observable Markov decision process (POMDP).
However, the POMDP-based strategies typically presuppose the availability of co-occurrence probabilities. This assumption is not always practical, and they struggle to adapt to new environments due to a lack of object-level commonsense knowledge.
Under the same setting, \cite{zsos} tries to infer the correlations between targets and known stationary items by learning from large image datasets, aiming for a general way to obtain correlation information. Nevertheless, it falls short of capturing the room-level spatial layout due to the limited perspective from the image form dataset, which is critical for adapting room-scale target search challenges.

Unlike previous approaches that fail to capture room-level and object-level knowledge for effective target localization, we introduce CSG-TL. Our approach combines both levels of knowledge, significantly improving the accuracy of target localization and resulting in better search performance.

\textbf{Scene graph modeling and inference.}
Scene graphs  \cite{sg1,sg2} abstract scenes into structured representations, mapping objects to nodes and their relationships to edges, streamlining complex tasks like image retrieval \cite{imgre} and visual question answering \cite{vqa}.
However, traditional robotics focuses more on building scene graphs during exploration \cite{hydra} than leveraging them to guide target localization. 
\cite{comsgnn} addresses this gap by developing spatial commonsense graphs (SCG), which model objects based on their spatial relationships within partially observed environments and integrate commonsense knowledge by connecting concept nodes from ConceptNet \cite{conceptnet} to relevant scene items node, trained on dataset to locate objects in unobserved areas.
However, this approach faces two main issues. Firstly, many indoor items share the same concept node provided by ConceptNet, which can lead to a fully connected graph that obscures the unique object-level commonsense attributed to each object.
Secondly, predicting the specific spatial coordinates of targets based solely on incomplete room-level knowledge from partially observed scenes results in heavy reliance on the training dataset. 
This can lead to confusion, such as expecting to predict the location of a bed just by giving a general perception of a living room. 

To address those limitations, we introduce a commonsense scene graph (CSG), a heterogeneous graph that incorporates
a complete room-level layout and object-level commonsense knowledge within its nodes and edges. This model prevents the lack of room-level information, which may cause unstable results. Besides, our object-level commonsense knowledge is obtained from an LLM that avoids the overlap from ConceptNet. This comprehensive model allows a more reasonable inference of target locations based on our rich correlations with both room-level and object-level knowledge.

\begin{figure*}[!htpb]
    \centering
    \includegraphics[width=1\linewidth]{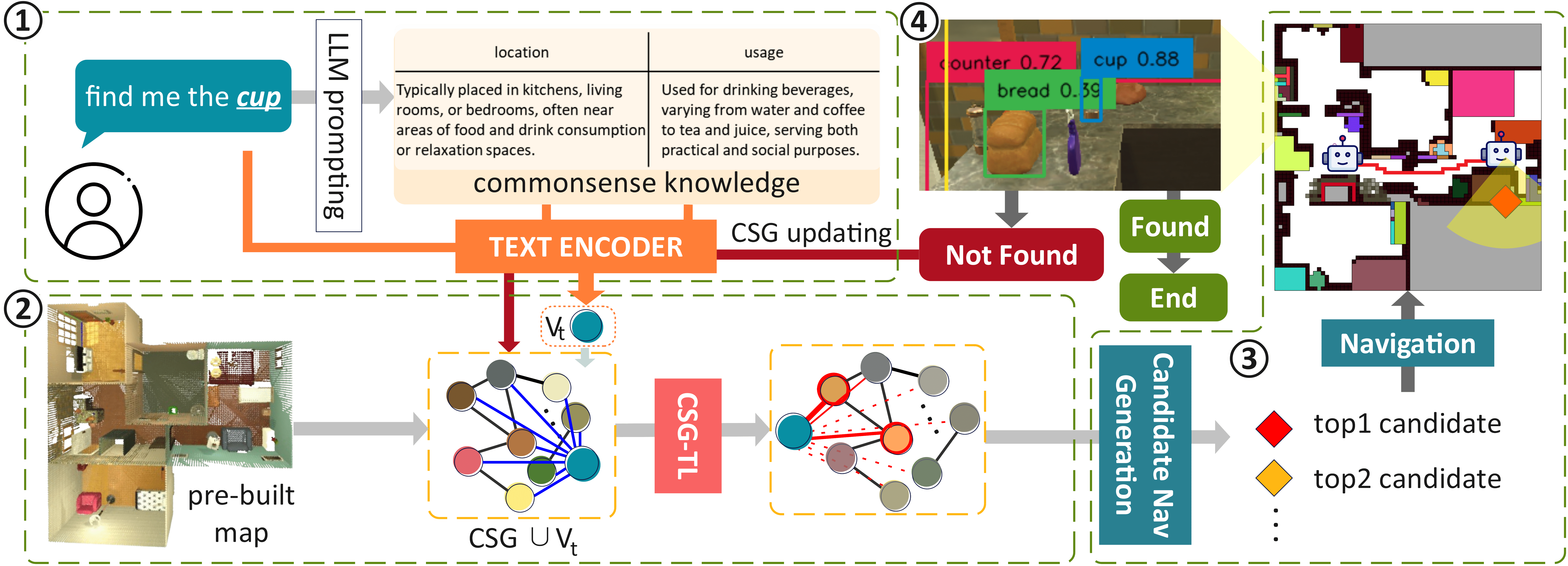}
    \caption{\textbf{CSG-based object search (CSG-OS) pipeline.} Firstly, the user queries the target object, which is then encoded with LLM-derived commonsense knowledge to form the target node $V_t$. Following this, the CSG is then constructed from a pre-built map of stationary items, incorporating $V_t$ for target localization through CSG-TL, detailed in Sec. \ref{sec:4-1}. Thirdly, nodes correlated with the target are clustered based on their locations and the likelihood of their predicted correlations, establishing a set of candidate search points. Finally, the robot navigates to the first candidate points in turn to search for the target. If found, the task is done successfully. Otherwise, the robot updates the CSG by newly detected objects and repeats the search steps until it finds the target or exceeds a preset threshold on thenumber of steps.}
    \label{fig:4-3}
\end{figure*}

\section{Problem Formulation}
In this work, we tackle the challenge of object search, explicitly focusing on scenarios relevant to household robots equipped with a pre-built map containing stationary items (e.g., furniture, fixtures). Mimicking the human search strategy, we propose the commonsense scene graph-based object search (CSG-OS) framework. 
This framework enhances the robot's search efficiency by initially predicting potential target locations based on their correlation at both room and object levels with stationary items (CSG-TL), and subsequently optimizing the search strategy by weighing the spread of these likely locations against the robot's present location. Fig. \ref{fig:4-3} shows the pipeline of our framework.

The commonsense scene graph (CSG) is initially constructed from a pre-built map $M$, with nodes $V = \{v_i | i \in  \{1,...,N\} \}$ representing the $N$ stationary items within the scene, $E = \{e_{ij} |i \in \{1,...,N\},   j \in \mathcal{N}_{i}\}$ representing the edge between node $V_i$ and $V_j$, where $\mathcal{N}_{i}$ denotes the set of neighbors for node $V_i$. Sec. \ref{sec:4-1} introduces how to build the CSG from a pre-built map, and the structure of the commonsense scene graph-based target localization (CSG-TL) model. Following this, Sec. \ref{sec:4-2} details the CSG-OS framework.

\section{Proposed Method}
This section provides a detailed discussion of the core component and the whole pipeline of our proposed CSG-OS framework. It includes the methodology for CSG construction, the generation of the dataset for training CSG-based target localization (CSG-TL), and the CSG-based object search (CSG-OS) framework.

\subsection{Commonsense Scene Graph\label{sec:4-1}} 
\textbf{Data Generation.}
Our dataset is derived from the ScanNet database \cite{scannet}, which contains 1,513 real-world scenes annotated with spatial relationships between objects. We use these scenes to extract room-level layouts and infuse object-level commonsense knowledge $C$ into the scene graph, using GPT-4 API for LLM-based queries. To construct the Commonsense Scene Graph (CSG), we categorize nodes into stationary items and daily movable objects, the latter being set as target objects for localization. For stationary items, which serve as known nodes in the CSG, we query descriptions of their actual locations and usage, forming the commonsense knowledge $C_{obj\_sta} = \{x_{loc}, x_{usage}\}$. For movable objects, which will later be trained as target objects, we similarly query their daily locations and usage to build $C_{obj\_tar}$. For edges, we query descriptions of spatial relationships and functional links between object pairs, generating the commonsense knowledge $C_{edge} = \{e_{geo}, e_{fun}\}$. The prompts used are similar to those in ConceptGraph \cite{conceptG}, and the details are provided in our code.

Similar to the previous scene graph setting \cite{zsos}, two objects are considered connected if the Euclidean distance between them is less than the threshold $d_{thre}$ or if they share a receptacle relationship (e.g., “on” or “inside”). Furthermore, if an object has no other object supporting the above condition, we identify the closest stationary item in the room to serve as its neighboring object. The edge label is created using the following rule:

\begin{equation}
E\left(V_{i}, V_{j}\right)=\left\{
\begin{array}{ll} \label{eq:1} 
1 & \left\|V_{i}-V_{j}\right\|<d_{\text {thre }} \\
1 & \text {receptacle}\left(V_{i}, V_{j}\right)  \\
1 & \text {if } V_{i} \text { has no close or receptacle } \\
  & \text {objects and } V_{j} \text { is nearest to } V_{i} \\
0 & \text {else} 
\end{array}\right.  
\end{equation}

\begin{figure}[H]
    \centering
    \includegraphics[width=0.95\linewidth]{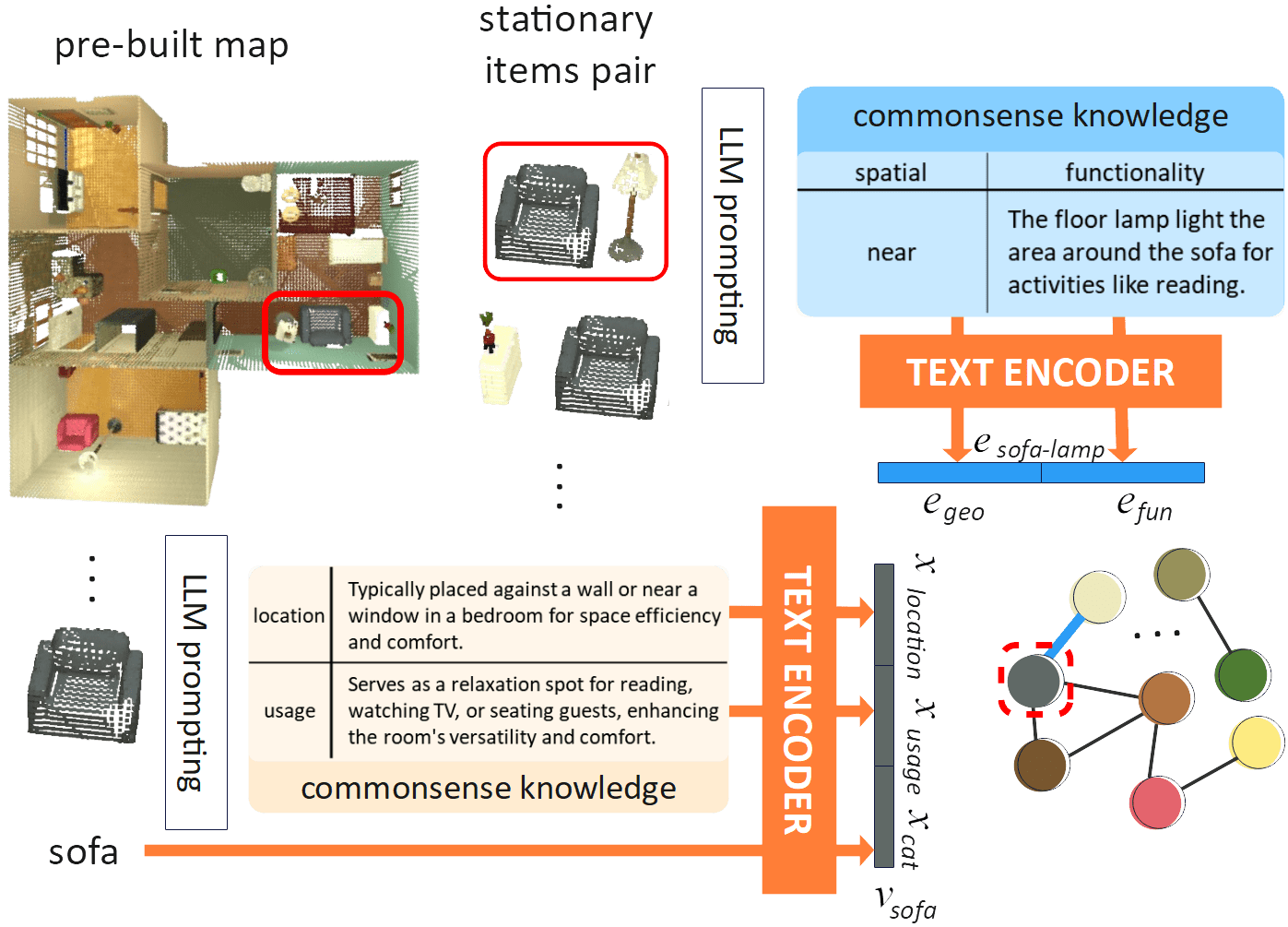}
    \caption{Illustration of the CSG construction process: The scene graph is initially created by identifying stationary items from the pre-built map based on spatial relationships, as defined in Eq. \ref{eq:1}. Subsequently, commonsense knowledge relevant to human inference of correlations is incorporated through LLM prompts. }
    \label{fig:sec4-1}
\end{figure}

According to those, the nodes feature $V=\{x_{cat}, x_{loc}, x_{usage}\}$ is the concation of the object commonsense knowledge $C_{obj} = \{x_{loc}, x_{usage} \}$ with the encoding of the object category $x_{cat}$, similarly, the edge feature $E = \{e_{geo}, e_{fun} \}$ is the commonsense knowledge $C_{edge}$. One example is presented in Fig. \ref{fig:sec4-1}.

\textbf{The Model of CSG-based Target Localization.}
To facilitate an efficient search process, we aim to identify regions most relevant to the target by utilizing the advanced inferential correlations provided by the commonsense scene graph (CSG) structure. With this objective, we present a target localization model CSG-TL, a key component of our CSG-OS framework.

\begin{figure}[htb]
    \centering
    \includegraphics[width=0.99\linewidth]{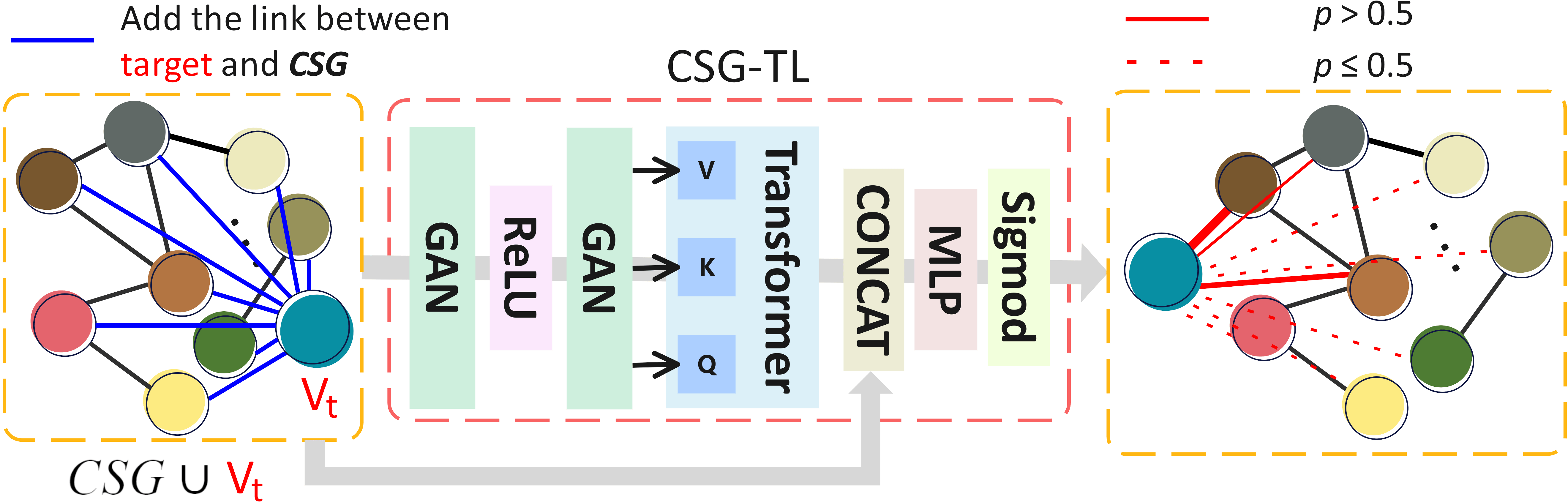}
    \caption{Illustration of the CSG-TL structure. Given the CSG and a target node $V_t$, the CSG-TL estimates the probability of correlation between the target and nodes within the CSG. A correlation exists if $p>0.5$ indicates a likely co-occurrence between the target and the respective object.}
    \label{fig:sec4-2}
\end{figure}

The CSG-based target localization (CSG-TL) model computes the likelihood of a target's correlation with objects in CSG using a multi-modal fusion module, which can be represented below:

\begin{equation}
\label{eq:4-2}
    p =\text{CSG-TL}(V_t, V, E)
\end{equation}

\noindent where $p$ is the likelihood between target $V_t$  and the other nodes $V = \{v_i|v \in \{1,...,N\}\}$ in CSG, and $E= \{e_{ij}|i,j \in \{1,...,N\}\}$ represents the edge in CSG. Fig. \ref{fig:sec4-2} details the structure of CSG-TL. Node features undergo a two-layer graph attention network (GAN) to fuse their feature $v_i''$ first, then pass a transformer by utilizing the edge feature as attention in Eq. \ref{eq:sec4-3}:

\begin{equation}
\label{eq:sec4-3}
    \alpha_{ij}=\text{ReLU}( (Q_i, K_j+e_{i j})/\sqrt{||K_j||} )
\end{equation}

\noindent where $Q_i, K_j$ are corresponding transformer values of node $V_i$ and its neighbors $V = \{V_j| j \in \mathcal{N}_{i}\}$, respectively.
The computed attention weights $\alpha$ are then forward used to fuse the node feature from  $v_i''$ to $v_i^*$, and finally concatenate the original and fused feature as $\hat{v} = [v*,v]$ of each node as the final feature. Finally, we use $[\hat{v}_t,\hat{v}_i]$, the concatenation of target node feature $\hat{v}_t$ and all others in CSG to represent the edge feature between them, using an MLP and a sigmoid to normalize the likelihood to $[0,1]$. If the likelihood exceeds $0.5$, it suggests the presence of an edge, indicating that the target is likely associated with the object; otherwise, it is not.

\textbf{Loss Function.}
The binary cross-entropy loss is utilized to measure the discrepancy between the predicted probability of a correlation link's existence, denoted by $p$ and the ground truth label $p_{gt}$ as:

\begin{equation}
\label{eq:4-4}
    \begin{array}{r}
\mathcal{L}=-\frac{1}{N} \sum_{i=1}^{N} p_{g t, i} \cdot \log \left(p_{i}\right)+ \\
\left(1-p_{g t, i}\right) \cdot \log \left(1-p_{i}\right)
\end{array}
\end{equation}

\noindent where $N$ is the number of total nodes in one graph. In our approach, we set 32 graphs as a batch and minimize the loss $\mathcal{L}$ for all graphs. Additionally, we operate under the assumption that the target will likely co-occur with nodes if their probability of correlation exceeds 0.5. 

\subsection{CSG-based Object Search\label{sec:4-2}}

Leveraging the target localization capabilities of the CSG-TL model, we introduce the CSG-based object search (CSG-OS) framework, designed to efficiently tackle the object search challenge. The overall CSG-OS pipeline is illustrated in Fig. \ref{fig:4-3}.

\textbf{Target Object Encoding and CSG-based Target Localization.}
Adapting the encoding strategy from Sec. \ref{sec:4-1}, the target is represented as a node $V_t$, encoded with commonsense knowledge via a prompt LLM. As in dataset and AI2THOR simulator, target category $x_{cat}$ is provided directly, allowing for straightforward encoding as $V_t=\{x_{cat}, x_{loc}, x_{usage}\}$. In practical, real-world scenarios, users can specify their search targets in free-form language $l_{tar-des}$, which typically encompasses the category $l_{cat}$ along with possible additional descriptors or hints, like a general location. Such inputs allow for extracting more detailed commonsense knowledge through LLM prompts, as demonstrated in the TABLE. \ref{tab:sec-41}. The target node is thus formulated as $V_t=\{l_{cat}, x_{loc}, x_{usage}\}$, accommodating both direct category mentions and more elaborate commonsense knowledge.

\begin{table}[hptb]
\begin{tabular}{M{1.6cm}|M{2.6cm}|M{3.4cm}N}
Target & Location & Commonsense Knowledge &\\[4pt] \hline
\multirow{2}{*}[-5ex]{Table} & \textbf{Living room}, placed near sofas& Central in social gatherings for  food and drinks. &\\[14pt]  \cline{2-3}
&  {\textbf{Kitchen}, placed in the center} & Dining space for eating meals, a prep area for food preparation, and a gathering spot for ... &\\[32pt]
\hline
\underline{Cup} in \textbf{Living room} (query by user as target) & \textbf{Living room}, on tables, or other surfaces near seating areas. & Holding beverages during leisure activities, social gatherings, or while relaxing and watching TV.&\\[32pt]  
\hline
Cup (generic)& On tables in the living room, kitchen, etc. & A common object in home environments, typically used for holding beverages. Its location often varies based ...  &\\[28pt]  

\end{tabular}
\vspace{0.1cm}
\caption{Commonsense knowledge for various inputs via LLM prompts. 
}
\label{tab:sec-41}
\vspace{-0.3cm}
\end{table}

After this, we connect the target node $V_t$ to CSG by linking it with all nodes $V$ in CSG to localize the target by CSG-TL.

\textbf{Candidate Navigation Point Generation and CSG Updating.}
The correlation likelihood \(h_i\) between the target and each object \(\text{Obj}_i\) is predicted by the CSG-TL, as detailed in Sec. \ref{sec:4-1}. 
However, the scattered arrangement of these objects does not inherently support an efficient search. 
To address this, we propose an approach that efficiently guides object search by utilizing the location \( \text{loc}_i =  [x_i, y_i] \) of each object $\text{Obj}_i$ and spreads the likelihood within a defined radius \( r \), decreasing with distance from the object center on the map \( M \), and represented as follows:

\begin{equation}
\label{eq:sec4-5}
    \begin{aligned}
M[x,y]_i & = h_i \cdot \max(0, r - \sqrt{(x - x_i)^2 + (y - y_i)^2}) \\
M[x,y] &= min(1, \frac{1}{K}\left[\sum_{i=1}^{K} M_{i}[x, y]\right] + w \cdot K)
\end{aligned} 
\end{equation}

\noindent where $M[x,y]_i$ indicates the influence of the object $i$ at position $[x,y]$ on the map, $M[x,y]$ is the cumulative likelihood, reflecting the average influence from all $K$ correlated objects with the target, and an award with $w$ by the number of correlated objects. To identify candidate navigation regions, we partition the map \(M\) into square sections \(R_i\) with a side length of $2r$ and accumulate the aggregated likelihoods within each to calculate \(W_{R_i}\):

\begin{equation}
\label{eq:sec4-6}
    W_{R_{i}}=\sum_{[x, y] \in R_{i}} M[x, y]
\end{equation}

Candidate regions $\{R_1, R_2,...\}$ are ordered by their weights $W_{R_{i}}$, and we choose their center as the candidate navigation position $\{p_1, p_2,...\}$. Besides this, the distance between robot $p_{robot} = [x_{robot},y_{robot}]$ and  candidate navigation position also should be considered, so we construct a cost function to balance both:

\begin{equation}
\label{eq:sec4-7}
\begin{aligned}
    C\left(p_{i}\right) & =\alpha \cdot\left(1-\frac{W_{R_{i}}}{\max \{ W_{R_{j}}\}}\right) \\
     & + \beta \cdot\left(\frac{D\left(p_{\text {robot}}, p_{n}\right)}{\max \{D\left(p_{\text {robot}}, p_{k}\right)\}}\right)
\end{aligned}
\end{equation}

\noindent where \(\alpha\) and \(\beta\) are weights to balance the importance of the candidate region and the distance to the robot, ensuring a normalized scale with \(\alpha + \beta = 1\). $D(,)$ caluculates the distance between two points in the map $M$ by $A^*$. 
Upon reaching the candidate search position, the robot looks for the target. If successful, the search concludes. Otherwise, the CSG undergoes an update with new non-stationary objects encountered during the search, repeating the cycle of target localization and navigation until success or the maximum number of steps is reached.

\section{EXPERIMENTS}
This section discusses the experimental validation of our proposed commonsense scene graph-based object search (CSG-OS) framework and its key component, commonsense scene graph-based target localization (CSG-TL). Both of them outperform current work. Finally, we present an overview of CSG-OS's real-world deployment, highlighting the efficiency and practicality of our method.


\begin{figure*}[!htpb]
    \centering
    \includegraphics[width=1\linewidth]{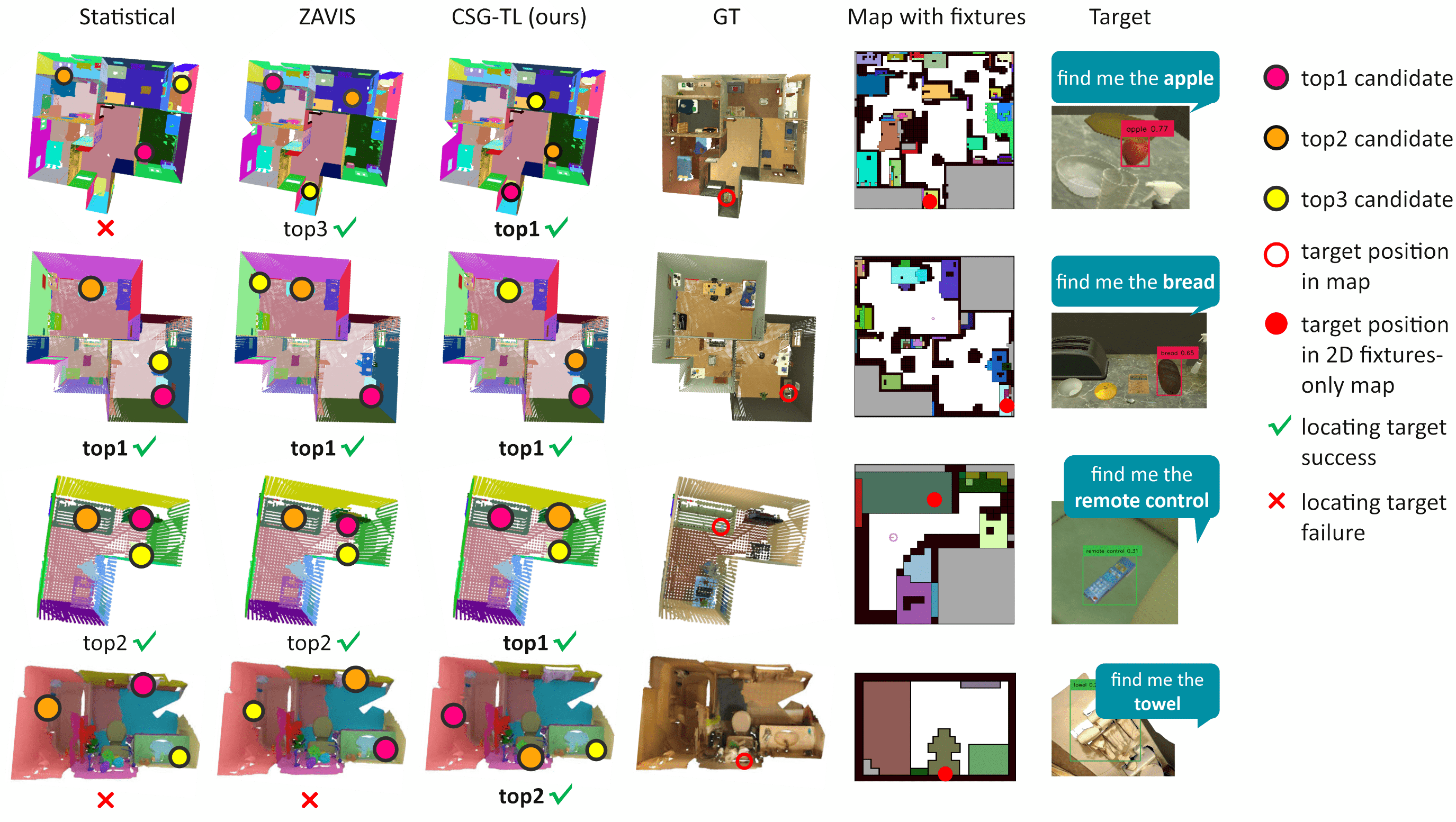}
    \caption{Visualization of target localization, the \tikzcircle[fill=magenta]{3pt} indicates the top1 candidate navigation point, \tikzcircle[fill=orange]{3pt} and \tikzcircle[fill=yellow]{3pt} representing the top2 and top3 candidates, respectively (calculation detailed in \ref{sec:4-2}). The prior knowledge they utilize is the pre-built map with stationary items only. The target's actual location in the full 3D map is marked with {hollow-red-circle}, and the corresponding location on the 2D fixtures-only map is denoted with {red-circle}. This visualization demonstrates the accuracy of our CSG-TL model in predicting the target's location against other methods. }
    \label{fig:5-1}
\end{figure*}


\begin{table*}[!h] 
\centering 
\begin{tabular}{ccccccN}
\hline
Link Pred. Acc         & Statistical    &  ZAVIS\cite{zavis}     &  ZSOS\cite{zsos}  & CSG-TL w/o commonsense knowledge     & CSG-TL       &\\[5pt] \hline
ScanNet                &  27.56\%       &   35.63\%   &  /         & {80.03\%}   & \textbf{89.73\%}     &\\[5pt]
AI2-THOR single room   &  47.32\%       &   59.25\%   &  67.22\%   & {73.11\%}   & \textbf{81.09\%}   &\\[5pt]  
AI2-THOR multi rooms   &  31.77\%       &   43.57\%   &  /         &  {65.21\%}   &  \textbf{78.21\%}   &\\[5pt] \hline
Commonsense knowlegde  & statistics   & statistics/Comet & category similarity & category similarity & category similarity/LLM  &\\[5pt] 
Method  & statistical   & statistical & graph & graph & graph  &\\[5pt]  \hline

\end{tabular}
\vspace{0.1cm}
\captionsetup{justification=raggedright,singlelinecheck=false}
\caption{Comparison of Link Prediction Accuracy across methodologies on the ScanNet dataset and AI2THOR environments.}
\label{tab:res5-1}
\vspace{-0.4cm}
\end{table*}

\subsection{The Result of CSG-TL}
As described in Sec. \ref{sec:4-1}, we train the proposed CSG-based target localization (CSG-TL) by ScanNet \cite{scannet}. 
We evaluate the trained CSG-TL model's link prediction performance using the ScanNet test split and the AI2-Thor environment. 
We consider 94 unique object categories in AI2THOR for target identification, consistent with previous studies\cite{zsos}.   
Objects are deemed to be connected if they meet the criteria outlined in Eq. \ref{eq:1}, with a distance threshold $d_{thre} = 1$, mirroring the parameters used in earlier studies \cite{zsos}.

In our evaluation, we compared our approach against three established methods:
\begin{itemize}
  \item \textbf{Statistics-based method:} Utilizing statistical correlations between targets and stationary items from the dataset, considering stationary items are linked with the target if their statistical correlation exceeds $0.5$.
  \item \textbf{Correlation via Knowledge Graph \cite{zavis}:} Enhancing statistical correlations with semantic connections from ConceptNet directly applies to visual perception, providing additional contextual insights.
  \item \textbf{GNN model trained on Visual Genome dataset \cite{zsos}:} Leveraging GNN to predict object correlations based on the Visual Genome image dataset.
\end{itemize}

To calculate accuracy, we first define a binary indicator function $H(x)$, which distinguishes connections based on a threshold of 0.5 for correlation between the target and other objects. Accuracy is then calculated by averaging the outcomes across all scenes within the dataset $G$, as shown in Eq. \ref{eq:5-1} below:

{\footnotesize
\label{eq:5-1}
\begin{align}
H(x) &= \begin{cases}
1 & \text{if } x \geq 0.5 \\
0 & \text{if } x < 0.5
\end{cases} \\
\text{Acc} &= \frac{1}{G} \sum_{g = 1}^{G}\left(\frac{\sum_{i = 1}^{N}H( \text{CSG}[V_{\text{target}}, V_{i}]) \wedge GT[V_{\text{target}}, V_{i}]}{\text{Num}[V_{\text{target}}, V_{i}]}\right)  
\end{align}
}

The performance results are presented in Table. \ref{tab:res5-1}. showcase the superior accuracy of our CSG-TL model over other methods within both the ScanNet and AI2THOR environments, demonstrating the significant advantage of our proposed CSG, integrating object-level commonsense knowledge via LLM into the scene graph built by room-level layout. As indicated in Table. \ref{tab:res5-1}, the method bases purely on statistical correlations performs the least effectively. However, integrating semantic and commonsense insights to visual perception \cite{zavis} significantly improves accuracy, although it still falls short of the outcomes achieved by graph-based inference methods \cite{zsos} and ours. Notably, even the variant of our model without commonsense knowledge still surpasses the performance of ZSOS \cite{zsos} relying on training on partial scene views from images. Our complete CSG-TL model, enriched with commonsense knowledge, consistently demonstrates the highest performance, affirming that commonsense reasoning significantly enhances target localization capabilities.

We also provide a visual comparison of target localization across different scenarios in AI2THOR and ScanNet scenarios. Illustrated in Fig. \ref{fig:5-1}, our CSG-TL model pinpoints the most probable candidate positions for the target object.

\begin{table*}[ht!]
\centering
\begin{tabular}{l|llllllllllN}
\hline
             & \multicolumn{2}{l}{Kitchen} & \multicolumn{2}{l}{Bathroom} & \multicolumn{2}{l}{Bedroom} & \multicolumn{2}{l}{Living room} & \multicolumn{2}{l}{Multi rooms} &\\[5pt] 
Model        & SPL      & SR       & SPL      & SR       & SPL      & SR       & SPL      & SR       & SPL    & SR   &\\[5pt] \hline 
Greedy-NBV\cite{ZhengKicra2022}   & 11.61\%  & 31.03\%  & 14.34\%  & 34.48\%  & 16.92\%  & 26.67\%  & 7.13\%   & 20.00\%  & /        & /   &\\[5pt] \hline
COS-POMDP\cite{ZhengKicra2022}    & 20.45\%  & 41.38\%  & 30.64\%  & 55.17\%  & 24.76\%  & 40.00\%  & 24.99\%  & 43.33\%  & /        & /   &\\[5pt] \hline
Target-POMDP\cite{ZhengKicra2022} & 13.80\%  & 34.48\%  & 19.88\%  & 55.17\%  & 19.79\%  & 26.67\%  & 24.36\%  & 40.00\%  & /        & /   &\\[5pt] \hline
ZAVIS\cite{zavis}        & /        & /        & /        & /        & 30.17\%  & 81.42\%  & 19.69\%  & 75.00\%  & /        & / &\\[5pt] \hline
ZSOS\cite{zsos}         & 23.56\%  & 46.67\%  & 56.62\%  & \textbf{86.67\%}  & 28.91\%  & 53.33\%  & 35.01\%  & 59.25\%  & /        & /  &\\[5pt] \hline
CSG-OS (\textbf{ours})       & \textbf{42.92\%}  & \textbf{63.33\%}  & \textbf{67.11\%}  & \textbf{86.67\%}  & \textbf{59.79\%}  & \textbf{86.67\%}  & \textbf{41.36\%}  & \textbf{76.67\%}  & \textbf{43.67\%}  & \textbf{46.67\%}   &\\[5pt] \hline
\end{tabular}
\vspace{0.1cm}
\caption{Comparative results of search success rate (SR) and success path length (SPL) in AI2THOR simulations of different scenarios.}
\label{tab:sec-52}
\end{table*}

\begin{table}[H]
\centering
\begin{tabular}{ccccN}
\hline 
Scenario  & AI2THOR & ProcThor (multi-rooms) & Realworld     &\\[5pt] \hline
$w$ & 0.05 & 0.05  &  0.1   & \\[3pt]
$\alpha$ & 0.4 & 0.6   &  0.5    & \\[3pt]  
$\beta$ & 0.6 & 0.4  &  0.5  & \\[3pt]
\hline
\end{tabular}
\vspace{0.2cm}
\caption{Hyperparameters setting }
\label{tab:sec4-4}
\vspace{-0.4cm}
\end{table}

\subsection{Simulation Result}
In this section, we present the efficacy of the proposed commonsense scene graph-based object search (CSG-OS) framework, benchmarked within the AI2THOR \cite{ai2thor} simulation. The AI2THOR provides a diverse set of 120 scenarios spanning four distinct environments: kitchens, living rooms, bedrooms, and bathrooms. ProcTHOR \cite{ProcTHOR} also extends this offering with more than 10,000 simulated household scenes, accommodating single and multi-room layouts. They include hundreds of unique objects with different shapes and textures. The robot is capable of executing basic movement commands, including ${\text{MoveAhead}, \text{RotateLeft}, \text{RotateRight}}$. The $\text{MoveAhead}$ command propels the robot forward by $0.25m$, while $\text{RotateLeft}$ and $\text{RotateRight}$ commands turn the robot $45^{\circ}$ to the left or right, respectively.

All these methods aim to improve object search by exploring correlations between the target and known stationary items. They employ approaches ranging from statistical distributions \cite{ZhengKicra2022}, extracting categorical similarities from knowledge graphs \cite{zavis}, to leveraging learning models trained on image datasets like Visual Genome \cite{zsos}. Similar to our method, they also dynamically update the probability of correlation between the target and other recognized objects besides stationary items during the search, allowing for continuous refinement of search strategies.
All methods were given equal search steps and tested on identical target object classes to maintain comparative fairness. The YOLOv5 object detector \cite{yolov5} was implemented as the standard perception module for all methods to ensure consistency. A successful search is defined by the robot's detection of the target object within a $1.0m$ Euclidean distance. Based on these, we set the hyperparameters in Eq.\ref{eq:sec4-5} and Eq.\ref{eq:sec4-6} as TABLR \ref{tab:sec4-4}.

The experimental results are shown in Table \ref{tab:sec-52}, assessing the search tasks using the Success Rate (SR) and the Success weighted by (normalized inverse) Path Length (SPL).
These results demonstrate our proposed CSG-OS's superiority over existing methods in terms of success rate and search efficiency (SPL). Notably, the traditional statistical correlation approach \cite{ZhengKicra2022} registers the lowest success rates, while ZAVIS \cite{zavis} incorporates semantic insights from statistical correlation enhanced with knowledge graphs into visual perception, marking a noticeable improvement. ZSOS \cite{zsos}, which derives correlations from image datasets without relying on predefined statistical knowledge of the environment, demonstrates further progress but falls short of providing a comprehensive understanding of the entire house layout.
Our CSG-OS stands out, reaching the SOTA performance even without specific environmental knowledge beforehand. This achievement highlights the importance of incorporating commonsense knowledge into the scene graph, markedly enhancing its inferential capabilities.


\subsection{Real-world Implementation and Results}

\begin{figure}[htpb]
    \centering
    \includegraphics[width=0.95\linewidth]{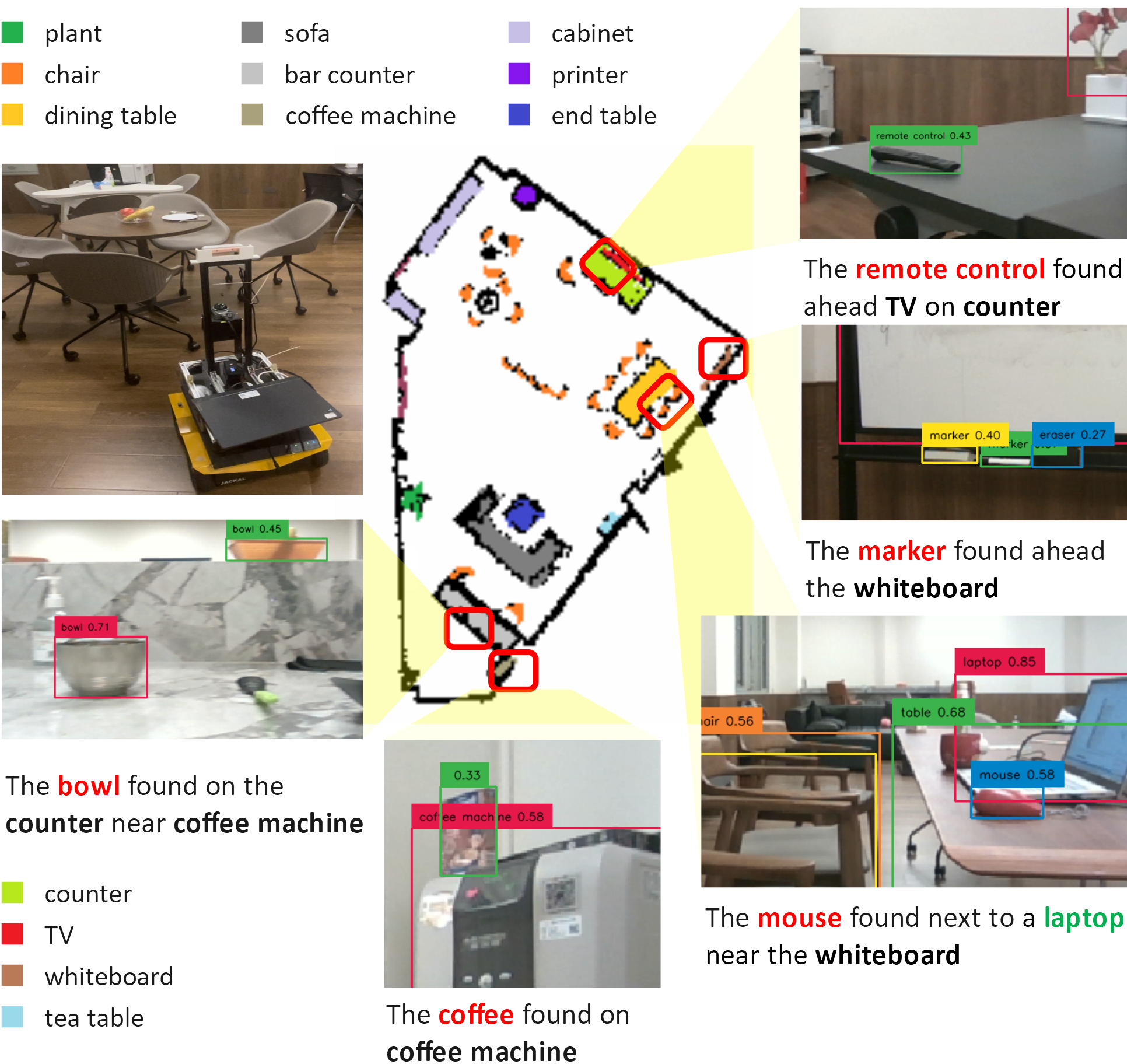}
    \caption{Real-world Implementation of CSG-OS on the Jackal Robot. The map incorporates stationary items utilized to construct the commonsense scene graph (CSG). Targets are highlighted in red, and are dynamically searched by the robot. Green denotes objects identified during the search that update the initial CSG, enhancing the system's predictive accuracy for target localization.}
    \label{fig:sec5-2}
\end{figure}

The practical effectiveness of the commonsense scene graph-based object search (CSG-OS) pipeline is further validated through deployment in a real-world setting. We employed a Jackal unmanned ground vehicle, equipped with a RealSense L515 camera for visual perception and a Livox Mid-360 LiDAR for precise localization. The robot autonomously explored the environment to construct a semantic map, exclusively recording and marking stationary items, as depicted in Fig. \ref{fig:sec5-2}. Leveraging this semantic map along with commonsense knowledge generated from the same prompted GPT-4 API queries described in Sec. \ref{sec:4-1}, we constructed a commonsense scene graph (CSG) to perform target localization, followed by subsequent object search tasks.

To evaluate our approach, we created a home-like environment where seven volunteers were instructed to place five types of commonly used objects in various natural locations. The Jackal robot was then tasked with locating and retrieving these items by utilizing the CSG-OS framework, effectively demonstrating its ability to perform real-world target localization and search.

\begin{table}[H]
\begin{tabular}{P{1.2cm}P{0.9cm}P{0.8cm}P{0.8cm}P{0.8cm}P{1.8cm}N}
\hline
Succ.            & bowl & coffee& mouse   & marker  & remote control   &\\[5pt] \hline
CSG-OS on Jackal & 100.00\% & 85.71\%  &  85.71\%  & 71.43\% & 71.43\%        & \\[12pt]  \hline
\end{tabular}
\vspace{0.1cm}
\caption{Real-world object localization success rates by CSG-OS }
\label{tab:sec5-3}
\vspace{-0.5cm}
\end{table}

The results of the Jackal robot’s performance in searching each object are presented in Table \ref{tab:sec5-3}, demonstrating the effectiveness of the CSG-OS in real-world applications. The task of \textit{bowl} searching achieved a $100\%$ success rate, attributable to its predictable placement on counters or tables and easy detection. The \textit{mouse} followed closely, benefiting from its strong association with laptops, which can be detected easily, the similar conclusion of \textit{coffee}. However, the success rates for the \textit{marker} and \textit{remote control} are relatively low. Although these items are often associated with whiteboards and TVs, they present greater challenges when displaced from their usual settings, and they are difficult to detect because of their small size and colors that easily blend with their surroundings.

\section{CONCLUSION}
In this study, we introduced an innovative Commonsense Scene Graph (CSG) that utilizes commonsense knowledge at both the room and object levels to improve scene modeling and target localization, mirroring human reasoning processes. Our approach markedly improves upon traditional object search strategies, which rely on correlation-based methods or learn from partial image views limited to a single type of knowledge. Our CSG-based object search (CSG-OS), facilitated by CSG-based target localization (CSG-TL), demonstrates effective zero-shot object search capabilities both in the AI2THOR simulator and real-world scenarios. 
While our framework demonstrates significant advancements, it encounters challenges in adapting to unconventional household environments like laboratories or certain public areas and struggles to interpret complex language-based goals within a scene. Our future work will aim to expand the adaptability of our approach to a wider range of scenarios, where language understanding will be pivotal in guiding robotic tasks effectively.


\end{document}